\documentclass[10pt,twocolumn,letterpaper]{article}

\usepackage{wacv}              %

\usepackage{graphicx}
\usepackage{amsmath}
\usepackage{amssymb}
\usepackage{booktabs}

\usepackage{amsmath,amsfonts,bm}

\def\eqref#1{equation~\ref{#1}}

\def\1{\bm{1}}

\DeclareMathAlphabet{\mathsfit}{\encodingdefault}{\sfdefault}{m}{sl}
\SetMathAlphabet{\mathsfit}{bold}{\encodingdefault}{\sfdefault}{bx}{n}

\usepackage{url}

\usepackage{graphicx, color}
\usepackage{tikz}
\usepackage{comment}
\usepackage{amsmath,amssymb} %
\usepackage{color}
\usepackage{ragged2e}
\usepackage{xcolor}         %
\usepackage{graphicx}
\usepackage{multirow}

\usepackage{subcaption}

\usepackage{caption}
\usepackage{float}

\usepackage{url}            %
\usepackage{booktabs}       %
\usepackage{amsmath}
\usepackage{amsfonts}       %

\usepackage{nicefrac}       %
\usepackage{microtype}      %
\usepackage{nicefrac}       %
\usepackage{wrapfig}

\usepackage{multirow}
\usepackage{makecell}
\usepackage{threeparttable}
\usepackage{sidecap}
\usepackage{caption}

\newcommand{\x}{\mathbf{x}}

\newcommand{\y}{\mathbf{y}}
\newcommand{\z}{\mathbf{z}}

\newcommand{\cc}{\mathbf{c}}
\newcommand{\g}{\mathbf{g}}
\newcommand{\btheta}{\boldsymbol{\theta}}
\newcommand{\bphi}{\boldsymbol{\phi}}

\usepackage{colortbl} %
\definecolor{mColor1}{rgb}{0.9,0.9,0.9}
\definecolor{mColor2}{rgb}{0.95,0.95,0.95}
\definecolor{non-photoblue}{rgb}{0.64, 0.87, 0.93}
\definecolor{lightblue}{rgb}{0.81, 0.94, 1.0}
\usepackage{tikz}
\usepackage{pifont}
\newcommand{\ymark}{\ding{51}}

\definecolor{mColor1}{rgb}{0.9,0.9,0.9}
\definecolor{mColor2}{rgb}{0.95,0.95,0.95}
\definecolor{non-photoblue}{rgb}{0.64, 0.87, 0.93}
\definecolor{lightblue}{rgb}{0.81, 0.94, 1.0}
\definecolor{lightorange}{rgb}{0.965, 0.835, 0.71}

\usepackage[textsize=tiny]{todonotes}

\usepackage[pagebackref,breaklinks,colorlinks]{hyperref}
\usepackage{orcidlink}

\usepackage[pagebackref,breaklinks,colorlinks]{hyperref}
\usepackage{algorithm}
\usepackage{algorithmic}
\usepackage{pifont}

\usepackage[capitalize]{cleveref}
\crefname{section}{Sec.}{Secs.}
\Crefname{section}{Section}{Sections}
\Crefname{table}{Table}{Tables}
\crefname{table}{Tab.}{Tabs.}

\def\wacvPaperID{295} %
\def\confName{WACV}
\def\confYear{2025}
\newcommand{\blue}[1]{\textcolor{black}{#1}}

\newcommand{\pink}[1]{\textcolor{black}{#1}}

\begin{document}

\title{GeneralizeFormer: Layer-Adaptive Model Generation across \\Test-Time Distribution Shifts}

\makeatletter
\renewcommand{\@fnsymbol}[1]{%
  \ifcase#1\or
    \ensuremath{\diamond}\or     %
    \textdagger\or
    \textdaggerdbl\or
    \textsection\or
    \textparagraph\or
    \textbardbl\or
    \textbullet\or
    \textdiv\or
    \textleaf\or
    \textmusicalnote\or
    \textmaltese\else
    \@ctrerr
  \fi}
\makeatother

\author{
  Sameer Ambekar\textsuperscript{1}\thanks{Currently with TU Munich, Germany}~,
  Zehao Xiao\textsuperscript{1},
  Xiantong Zhen\textsuperscript{1,2}\thanks{Currently with United Imaging Healthcare, Co., Ltd., China.}~,
  Cees G. M. Snoek\textsuperscript{1} \\
  \textsuperscript{1}AIM Lab, University of Amsterdam~
  \textsuperscript{2}Core42
}

\maketitle
\begin{abstract}
We consider the problem of test-time domain generalization, where a model is trained on several source domains and adjusted on target domains never seen during training. Different from the common methods that fine-tune the model or adjust the classifier parameters online, we propose to generate multiple layer parameters on the fly during inference by a lightweight meta-learned transformer, which we call \textit{GeneralizeFormer}. The layer-wise parameters are generated per target batch without fine-tuning or online adjustment. By doing so, our method is more effective in dynamic scenarios with multiple target distributions and also avoids forgetting valuable source distribution characteristics. Moreover, by considering layer-wise gradients, the proposed method adapts itself to various distribution shifts. To reduce the computational and time cost, we fix the convolutional parameters while only generating parameters of the Batch Normalization layers and the linear classifier. Experiments on six widely used domain generalization datasets demonstrate the benefits and abilities of the proposed method to efficiently handle various distribution shifts, generalize in dynamic scenarios, and avoid forgetting. Our code is available: \url{https://github.com/ambekarsameer96/generalizeformer}

\end{abstract}

\section{Introduction}

To counter the performance degradation of deep learning algorithms when test data strays from the training data distribution, the framework of test-time adaptation emerged \cite{wang2021tent, iwasawa2021test, zhang2023adanpc}. In test-time adaptation \cite{xiao2024beyond}, one trains a model only on source domains and adapts the source-trained model during inference on (unseen) target data. Various methods are proposed for adaptation to a single source distribution, as typical for image corruption \cite{sun2020test, wang2021tent, lim2023ttn}, and for multiple source distributions, as common in domain generalization \cite{iwasawa2021test, xiao2022learning, zhang2023adanpc}. In this paper, we focus on the latter scenario, referring to it as test-time domain generalization.

\begin{figure}[t!]
\centering
\includegraphics[width=\linewidth]{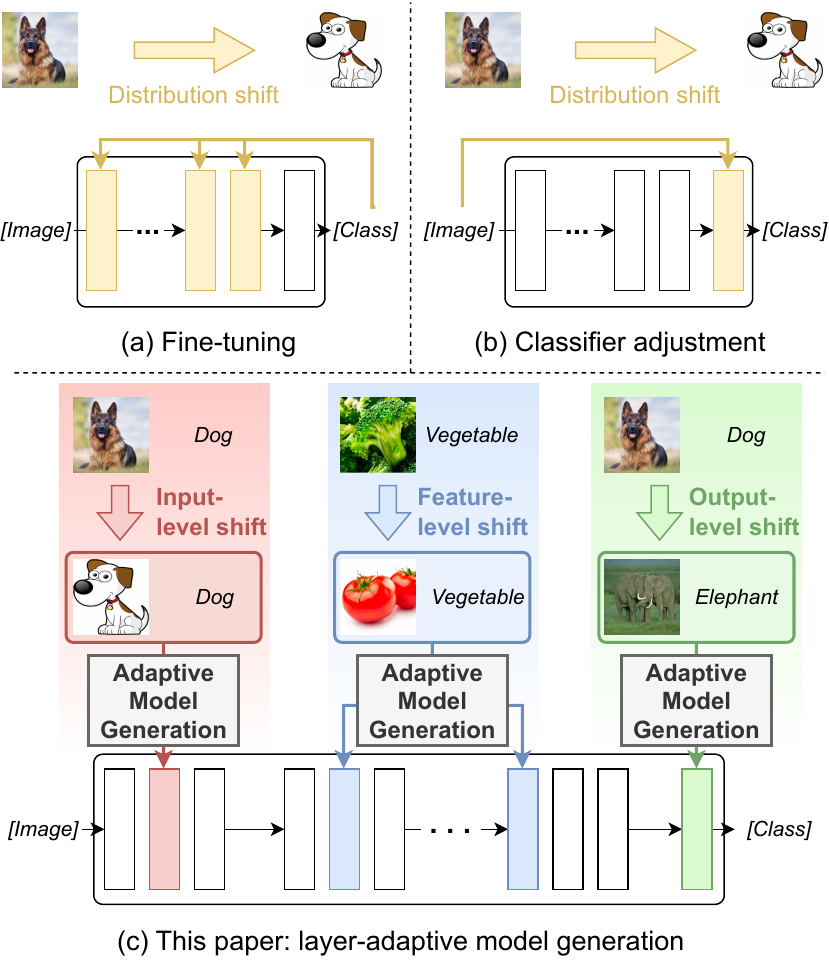}
\caption{\textbf{Illustration of test-time generalization methods.} (a) Fine-tuning methods update the model online with large batches of samples. 
(b) Classifier adjustment methods feedforwardly update the classifier according to the test data. 
(c) Our method adaptively generates target-specific parameters in different layers according to the target distribution, enabling it to handle various distribution shifts without fine-tuning. 
}
\label{fig1:all}
\vspace{-4mm}
\end{figure}

Current test-time generalization methods focus on fine-tuning the model with unsupervised loss functions on the target data \cite{wang2021tent,goyal2022test} or adjusting the linear classifiers during inference \cite{iwasawa2021test, zhang2023adanpc}. 
However, \blue{these methods utilize batches of target data at the same time \cite{liang2020we, kundu2022balancing} or in an online manner \cite{wang2021tent}.}
The alternative of classifier adjustment only considers the distribution shifts during classification, but ignores shifts resulting from feature extraction \cite{zhen_learning, lee2022surgical}. 
Moreover, these methods rely on online updates of model parameters on each specific target distribution and tend to forget the discriminative ability of previous source distributions, which makes them struggle in complex or dynamic scenarios \cite{yuan2023robust, zhang2023adanpc}.
Recently, Lee \etal \cite{lee2022surgical} propose to handle different distribution shifts by \emph{manually} selecting and fine-tuning layers, which, however, still requires different selection and fine-tuning operations for different specific target distributions.
Building on these problems and observations, we propose to adaptively generate target-specific parameters in different model layers to handle various distribution shifts without re-training the parameters (see also Fig. \ref{fig1:all}).

We propose \textit{GeneralizeFormer}, a transformer that generates target-specific layer-wise model parameters for each target batch at test-time. 
The target parameters are generated from the source-trained parameters, the target features, and the layer-wise gradient information.
With the layer-wise generation of the target parameters, the method can adaptively handle distribution shifts in different feature levels without manually selecting and fine-tuning parameters.
To enable model generation ability across distribution shifts, we train the transformer under a meta-learning strategy. At test time, GeneralizeFormer achieves \blue{adaptation} without online optimization of the model, making it especially effective in dynamic scenarios that suffer from multiple target distributions. This also helps the model to preserve the source model information and avoid forgetting after model \blue{adaptation}. To reduce the computational and time costs for \blue{adaptation}, we fix the parameters of the convolutional layers while generating the parameters of all Batch Normalization layers and that of the classifiers in the network.
Experiments on six domain generalization benchmarks and various distribution shifts show the ability of our method to handle source forgetting, dynamic scenarios, and multiple distribution shifts.

 \section{Related Work}

\noindent
\textbf{Domain adaptation and domain generalization.}
Distribution shifts have been widely investigated in domain adaptation \cite{long2015learning, wilson2020survey, you2019universal} and domain generalization methods \cite{arjovsky2019invariant, zhou2022domain, wang2022generalizing}.
Domain adaptation deals with the distribution shifts between training and test data by assuming the accessible labeled or unlabeled target samples during training \cite{ben2006analysis, farahani2021brief, csurka2017domain, guo2017calibration}. 
However, in real applications, the target data is usually unknown during training \cite{zhao2020domain}.
To solve this problem, domain generalization is proposed to train the model only on source domains and evaluate it on unseen target domains \cite{muandet2013domain, li2018domain, dou2019domain, gulrajani2020search, xiao2021bit, park2023test}. 
The problem of domain generalization is the lack of target information of the source-trained model, which may lead to the adaptivity gap \cite{dubey2021adaptive}.
Recently, source-free domain adaptation has emerged \cite{liang2020we, kundu2022balancing, wang2022exploring, roy2022uncertainty, fang2022source, jing2022variational, yu2023comprehensive, thopalli2023target} where source model is adapted to target domains at test time. 
Our test-time domain generalization setting is related to the above settings, where we train the model on several source domains as domain generalization. 
Rather than adapting the model to the entire target domain, we achieve \blue{adaptation} at each test step.

\noindent
\textbf{Test-time adaptation and generalization.}
Test-time adaptation is also an emerging method for handling distribution shifts at test time \cite{sun2020test, wang2021tent, liu2021ttt++, gao2022back, niu2023towards, sun2022dynamic, liu2024cross, yi2023source, zhang2023domainadaptor}. These methods usually train the model on a single source distribution and achieve adaptation at test time in an online manner, focusing on covariate shifts such as corruptions  \cite{wang2021tent, goyal2022test, zhang2021memo, niu2022efficient}.
The idea has also been applied to the domain generalization setting, where multiple source domains are available during training  \cite{du2020metanorm, li2018learning, xiao2023energy, jang2022test, zhang2023adanpc, ambekar2024variational, ambekar2023variational_iclr}. 
We refer to this setting as test-time domain generalization.
One commonly used test-time method is fine-tuning the source-trained model with an unsupervised loss like entropy minimization \cite{wang2021tent, zhang2021memo, zhou2021bayesian, goyal2022test, zhang2023domainadaptor,niu2022efficient}. As an alternative, changing the statistics of the Batch Normalization layers is also widely utilized \cite{schneider2020improving, du2020metanorm, you2021test, hu2021mixnorm, gong2022note,lim2023ttn}.
Another method for test-time adaptation and generalization is classifier adjustment \cite{iwasawa2021test, xiao2022learning, zhang2023adanpc}, where the prototypes or the last fully-connected layer of the model is adjusted at test time, without fine-tuning.
Our method focuses on the test-time domain generalization setting.
Lee \etal \cite{lee2022surgical} showed that fine-tuning different layers benefits different distribution shifts.
In contrast to the previous methods, we propose to adaptively generate different layers parameters according to each batch of target samples, which avoids fine-tuning and adaptively addresses different distribution shifts.

\noindent
\textbf{Model generation networks.} 
The idea of using neural networks to generate model weights was introduced by Hypernetworks \cite{ha2016hypernetworks}. 
This approach is usually used for compressing large models into smaller ones. Recently, the concept of hypernetworks to generate compressed networks has been explored by \cite{zhmoginov2022hypertransformer} to generate weights for models using task descriptions. 
The techniques are also utilized in natural language processing tasks \cite{pilault2020conditionally, mahabadi2021parameter} and \blue{neural representations~\cite{ashkenazi2022nern}.}
Different from these methods, we utilize transformers to generate the layer-wise target model parameters according to the source model, target features, and gradient information. The model generation is meta-learned to generalize to unseen target during source training.

\section{Methodology}

\noindent
\textbf{Preliminary.} Test-time domain generalization aims to generalize the model $\btheta_{s}$ trained on the source domains $\mathcal{S}$ to the unseen target domain $\mathcal{T}$, where $\mathcal{S}$ usually consists of several source domains $\{ D_s \}_{s=1}^{S}$. Naturally, $\mathcal{T}$ may also consist of several target domains $\{ D_t \}_{t=1}^{T}$. Here, $(\x_s, \y_s)$ and $(\x_t, \y_t)$ denote the data-label pairs on the source domain $D_s$ and target domain $D_t$, respectively. The objective of test-time domain generalization is to maximize the log-likelihood of the target data $p(D_t|\btheta_s)$, i.e., $p(\y_t|\x_t, \btheta_s)$.

Since distribution shifts between the source and target domains are expected, the source-trained model $\btheta_s$ usually struggles on the unseen target domain $D_t$, leading to unreliable predictions \cite{dubey2021adaptive}. In this case, the source model needs to be further generalized to the target domains at test time. One commonly used method for test-time domain generalization is fine-tuning \cite{wang2021tent, jang2022test}, which generalizes the source-trained model $\btheta_s$ to $\btheta_t$ before predicting on target data. The log-likelihood of the target data is then formulated as:
\begin{equation}
\label{tta} %
\begin{aligned}
    p(\y_t|\x_t, \btheta_{s}) & = \int p(\y_t|\x_t, \btheta_{t}) p(\btheta_{t}|\x_t, \btheta_{s}) d \btheta_{t} \\ 
    & \approx p(\y_t|\x_t, \btheta^*_{t}),
\end{aligned}
\end{equation}
where the integration of distribution $p(\btheta_t)$ is usually approximated by the maximum a posteriori (MAP) estimation. The MAP model $\btheta^*_{t}$ is obtained by fine-tuning with an unsupervised learning loss function like entropy minimization \cite{wang2021tent} or pseudo labeling \cite{jang2022test}.
However, by updating the model parameters through backpropagation, fine-tuning methods require expensive time and computational costs for generalization in each test step. 
Moreover, in real-world applications, the target domains are usually unpredictable, which means target samples from any domain will arrive at any test time step during the online inference and generalization \cite{wang2022continual, yuan2023robust, niu2023towards}. In these dynamic scenarios with continually changing distributions, the fine-tuning-based methods would lead to underfitting \blue{adaptation} due to limited target information and distribution forgetting caused by loss of source information.
To deal with these problems, we propose to directly generate the target-specific model $p(\btheta_{t}|\x_t, \btheta_{s})$ with a designed transformer network rather than obtaining it by MAP approximation through fine-tuning.

\begin{figure*}[t]
\centering
\vspace{-2mm}
\includegraphics[width=0.90\linewidth]{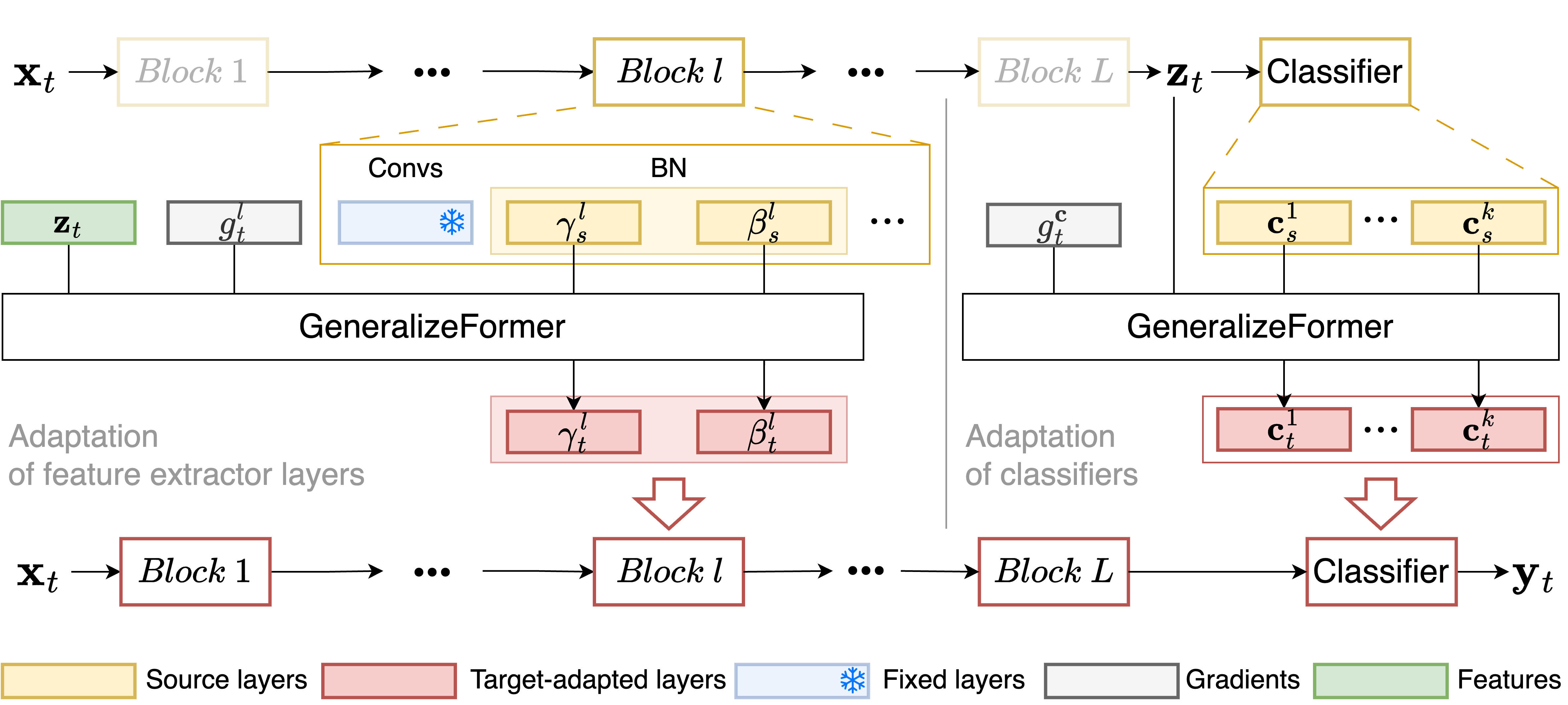}
\vspace{-2mm}
\caption{\textbf{Illustration of GeneralizeFormer.} 
We generate the model parameters of the classifiers and the Batch Normalization layers at different levels with GeneralizeFormer. 
The GeneralizeFormer takes the source-trained parameters, target features, and layer-wise gradients as input and outputs the target-specific parameters.
By considering the layer-wise information, the method adaptively generates target parameters for different levels of layers, enabling the model to handle various distribution shifts.}

\label{fig:generalformer}
\vspace{-4mm}
\end{figure*}

\noindent
\textbf{GeneralizeFormer.}
To achieve test-time generalization by directly inferring $p(\btheta_{t}|\x_t, \btheta_{s})$, we generate the target model $p(\btheta_t)$ with a transformer $\bphi$.
The log-likelihood in eq. (\ref{tta}) is:
\begin{equation}
\label{tta_gf} %
\begin{aligned}
    p(\y_t|\x_t, \btheta_{s}) & = \int p(\y_t|\x_t, \btheta_{t}) p_{\bphi}(\btheta_{t}|\x_t, \btheta_{s}) d \btheta_{t},
\end{aligned}
\end{equation}
where the generation of $p_{\bphi}(\btheta_t)$ is conditioned on the source trained model $\btheta_s$ and the target data $\x_t$.
Specifically, the information on the target data is necessary to generate the target data-specific model. 
In test-time generalization, the only available target information is the current batch of test samples $\x_t$ or just a single sample \cite{xiao2022learning}. We use the feature representations of the current target samples to integrate the target information for each test time step. The features are obtained by the source-trained model, i.e. $\z_t {=} f_{\btheta_s}(\x_t)$. 

Since our goal is to achieve good performance on the target data \blue{during generalization}, the generated target model needs to have good ability for feature extraction and classification.
Although the source-trained model parameters suffer from distribution shifts, the basic feature extraction or classification ability is necessary to achieve fast generalization with the limited number of target samples at test time. Therefore, we also utilize the source-trained parameters $\btheta_s$ as inputs for our GeneralizeFormer.

In addition to the target features from $\x_t$ and source-trained model parameters $\btheta_s$, we also include layer-wise gradient information of the model on the target samples. 
The gradients are obtained by $\g_t^l {=} {\partial \mathcal{L} (\x_t)} / {\partial \btheta^l_s}$, where $l$ denotes the model layers. The loss function $\mathcal{L} (\x_t)$ can be any unsupervised loss. Here, we use the common 
entropy minimization \cite{wang2021tent, zhang2021memo}.
By feeding the layer-wise gradient information into the transformer $\bphi$, the generated model $\btheta_t$ further considers the effects of different layers on the target samples. With the gradients $\g_t^l$ that are both data-specific and layer-specific, $\btheta_t$ not only contains more diverse target information but also achieves an adaptive layer-specific adjustment according to the source model and target data.

With the above inputs, the generation of the target-specific model $p_{\bphi} (\btheta_t)$ is achieved by:
\begin{equation}
p_{\bphi} (\btheta_t | \btheta_s, \x_t) = \{ p_{\bphi} (\btheta_t^l | \btheta_s^l, \x_t) \}_{l=1}^L = \{ \bphi (\btheta^l_s, \z_t, \g^l_t) \}_{l=1}^L,
\end{equation}
where $l {=} \{ 1, 2, \cdots, L \} $ denotes the layer of the model. Fig. \ref{fig:generalformer} illustrates the generation process.

\noindent
\textbf{Meta GeneralizeFormer.} To learn the ability of model generation across distributions, we further train the GeneralizeFormer under a meta-learning framework \cite{du2020learning, xiao2022learning}.
Specifically, we simulate distribution shifts during training by splitting the source domains $\mathcal{S}$ into meta-source domains $\mathcal{S}'$ and a meta-target domain $\mathcal{T}'$. The meta-target domain is selected randomly in each iteration to mimic diverse domain shifts. 

Per iteration, we train the meta-source model $\btheta_{s'}$ by minimizing the cross-entropy loss on the meta-source data by:
\begin{equation}
\label{metasource}
   \btheta_{s'} = \mathop{\min}\limits_{\btheta} \mathbb{E}_{(\x_{s'}, \y_{s'}) \in \mathcal{S}'} [\mathcal{L}_{\mathrm{CE}}(\x_{s'}, \y_{s'}; \btheta)]. 
\end{equation}
To mimic the test-time generalization procedure and learn the ability of model generation across distributions, we then train the transformer $\bphi$ by the meta-source model $\btheta_{s'}$ and the meta-target data $\x_{t'}$. Specifically, we obtain the target features $\z_{t'}$ and gradients $\g_{t'}^l{=}{\partial \mathcal{L} (\x_t)} / {\partial \btheta_{s'}^l}$ using the meta-target data. The meta-target model is then generated by $\btheta^l_{t'} {=} \bphi(\btheta^l_{s'}, \z_{t'}, \g_{t'}^l), \forall l=1,2,\cdots, L$. 

Since we have access to the actual meta-target labels $\y_{t'}$ during source training, we train the transformer by supervising the model generation procedure:
\begin{equation}
\label{metatar}
\bphi = \min _{\bphi} \mathbb{E}_{(\x_{t'}, \y_{t'}) \in \mathcal{T}' }[\mathcal{L}_{\mathrm{CE}}(\x_{t'}, \y_{t'}; \btheta_{t'})],
\end{equation}
where $\btheta_{t'} = \{ \btheta^l_{t'} \}_{l=1}^L$ denotes the generated parameters of different layers and $\mathcal{L}_{\mathrm{CE}}$ denotes the supervised cross-entropy loss. Intuitively, after multiple iterations, the cross-entropy on meta-source model $\btheta_{s'}$ in eq.~(\ref{metasource}) trains the basic feature extraction and classification ability of the model. The transformer network $\bphi$ further learns to generate specific model parameters across distributions based on the source-trained model and target data by the supervision in eq.~(\ref{metatar}).

At test time, the method directly generates the target-specific model by $\btheta^l_{t} {=} \bphi(\btheta^l_{s}, \z_{t}, \g_{t}^l), \forall l=1,2,\cdots, L$ for each batch of target data, without any fine-tuning operation and avoiding the unstable \blue{adaptation} or error accumulation in the dynamic scenarios.

\noindent
\textbf{Generating batch norm and classifier.} Due to the relationship between the source and target data, source model parameters of different layers may have different effects on the target samples \cite{lee2022surgical}.
Therefore, it is more effective to implement specific adjustments on various layers of parameters to handle different distribution shifts or different target samples. We achieve layer-specific adaptation by generating the model parameters of different layers and considering the specific gradient information.

To do so, we fix the large amount of parameters from the convolutional layers and generate parameters for the Batch Normalization layers at different levels and the classifiers as shown in Fig. \ref{fig:generalformer}.
In this case, the method is computationally more efficient since the batch norm affine and classifier layers are low dimensional with much fewer parameters than the convolutional layers.
Moreover, the normalization layers and classifiers are known to influence distribution shifts \cite{huang2017arbitrary,xiao2022learning,lim2023ttn,iwasawa2021test}. 
By adaptively generating the affine parameters for required normalization layers, our method adjusts the feature statistics at different levels of the source model, which addresses diverse types of distribution shifts.
The generation of the classifier parameters further handles the domain shifts at the semantic level. 
Overall, by generating BN and classifier parameters, we handle domain shifts across different feature levels efficiently.

Specifically, the Batch Normalization layers \cite{ioffe2015batch} first normalize the feature representations by $\bar{\z}_t {=} {\z_t - \mu} / {\sigma}$, where the statistics $\mu$ and $\sigma^2$ are obtained by the moving average of the training statistics. The normalized features are then scaled and shifted by $\hat{\z}_t {=} \gamma_s \bar{\z}_t + \beta_s$, where the affine parameters $\gamma_s$ and $\beta_s$ are learned during the source training. 
To achieve adaptation on different levels of layers, we generate the target data-specific affine parameters $\gamma_t$ and $\beta_t$ of all Batch Normalization layers by the transformer, with inputs of the source-trained affine parameters, target features, and the corresponding gradients. 
Different from previous methods \cite{wang2021tent} that update both affine parameters and statistics, our method only generates the affine parameters while maintaining the source statistics, which reduces the influence of small batch sizes at test time.

To generate the classifier parameters, we treat the source-trained classifier $ \cc^k_s $ of each category $ k $ as an input vector of the transformer and generate the target classifiers $ \cc^1_t, \cdots, \cc^K_t $ together by considering the target features and gradients information as shown in Fig. \ref{fig:generalformer}.

In summary, the generation of Batch Normalization affine parameters and classifier parameters are formulated as:
\begin{equation}
\begin{aligned}
& \gamma^l_t, \beta^l_t = \bphi (\gamma^l_s, \beta^l_s, \z_t, \g^{\gamma^l}_t, \g^{\beta^l}_t), \\
& \cc^1_t, \cdots, \cc^K_t = \bphi (\cc^1_s, \cdots, \cc^K_s, \z_t, \g^{\cc}_t),
\label{eqn_genbn}
\end{aligned}
\end{equation}
where $\gamma^l, \beta^l$ denote the affine parameters of the $l$\textit{-th} Batch Normalization layers and $\cc^k, k \in \{ 1, \cdots, K, \}$ denote the classifier of $k$\textit{-th} category.  
From now on, to ease notation, we use $\btheta$ to summarize the affine parameters $\gamma$ and $\beta$ of different Batch Normalization layers and classifier parameters $\cc$.

With the trained transformer model $\bphi$, our method adaptively generates the parameters of different Batch Normalization layers and classifiers for different distribution shifts at test time.
Since we fix the convolutional layers and generate other parameters by feedforward calculation of the transformer without online optimization, the model adjustment is more efficient than online fine-tuning methods \cite{wang2021tent, iwasawa2021test}.
Note that we generate Batch Normalization and classifier parameters in this work, but the method can also be extended to any linear layer, which benefits other model architectures.

\noindent \pink{\textbf{Rationale behind utilizing transformer. }} \pink{We introduce a transformer for parameter generation, whose attention mechanism effectively aggregates useful knowledge in source parameters and target features to avoid information loss.
We further consider layer-wise gradients per target batch as input of the transformer, which indicates the relationships between each layer of the source parameters and each target batch. By doing so, the gradients guide model generation for each layer and for different target domains, batches, and even samples.
This reduces error transmission among target samples and layers while enhancing the generalization ability across samples and domain shifts.}

\section{Experiments}

\subsection{Datasets and implementation details}

\vspace{1mm}
\noindent
\textbf{Six datasets.} 
We perform the experiments on image classification problems and demonstrate its effectiveness on six datasets namely: \textit{PACS} \cite{li2017deeper}, \textit{VLCS} \cite{fang2013video}, \textit{Office-Home}\cite{venkateswara2017deep}, \textit{TerraIncognita} \cite{beery2018recognition}, \textit{Living-17} \cite{santurkar2020breeds}, \textit{Rotated MNIST} and \textit{Fashion MNIST} \cite{piratla2020efficient} and provide details in supplementary.

\vspace{2mm}
\noindent
\textbf{Implementation details.}
We evaluate on the online test-time domain generalization setting defined by Iwasawa and Matsuo \cite{iwasawa2021test}, where the data at test-time is iteratively incremented and we update the model per target batch externally through our $\bphi$ without backpropagation of the source trained model. 
The ERM baseline refers to directly evaluating the source-trained model on the target set \cite{gulrajani2020search}. 
\textcolor{black}{We implement the lightweight $\bphi$ model by the PyTorch transformer encoder module with 8 layers.
We also tried 4-layer and 2-layer transformers, which consume fewer memory with similar performance as shown in the supplemental material.}
\pink{For the transformer, we train it for 10,000 iterations alongside the ResNet. The source parameters, target features, and gradients are all formatted to the same dimension as the source parameters (e.g., 512 for the last block for ResNet-18) and then concatenated and utilized as input tokens to the transformer. 
The attention is calculated between these inputs to enhance each other according to their relationships.
We use the output features of the source parameters as the generated target parameters, which therefore match the dimensions of the source ones.}
For source training, we utilize a learning rate of $1e-4$ with Adam optimizer for the ResNet18 and ResNet-50 models (pre-trained on Imagenet) and the $\bphi$ model with a batch size of 64 during training. The hyperparameters for the best model during source training have been selected by the validation set following \cite{iwasawa2021test,gulrajani2020search}. \textcolor{black}{For test-time generalization, we utilize a small batch size of 20 samples. We will show in the ablations that our method also performs well with smaller batches. We generate the target model parameters by passing the inputs to the lightweight transformer encoder $\bphi$ without any backpropagation on the backbone model.} These new weights are directly assigned to the source-trained model. There are no additional hyperparameters involved. We run all experiments with five different random seeds. We have provided additional implementation details and our algorithm in the supplementary.

\begin{table*}[t]
\centering

\vspace{-3mm}
	\resizebox{1.95\columnwidth}{!}{%
		\setlength\tabcolsep{4pt} 
\begin{tabular}{lllllllll}
\toprule
& \multicolumn{2}{c}{\textbf{PACS}}     & \multicolumn{2}{c}{\textbf{VLCS}}    & \multicolumn{2}{c}{\textbf{Office-Home}}    & \multicolumn{2}{c}{\textbf{TerraIncognita}}\\ 
\cmidrule(lr){2-3} \cmidrule(lr){4-5} \cmidrule(lr){6-7} \cmidrule(lr){8-9}
\quad \textbf{Methods} & ResNet-18 & ResNet-50 & ResNet-18 & ResNet-50 & ResNet-18 & ResNet-50 & ResNet-18 & ResNet-50 \\ \midrule
\quad ERM baseline & 79.6 & 85.7 & 75.8 & 77.4 & 61.0 & 67.5 & 35.8 & 47.2 \\
\midrule
\multicolumn{9}{l}{\textit{Standard domain generalization}} \\
\quad Arjovsky \etal \cite{arjovsky2019invariant} & \textcolor{black}{80.9} & 83.5 & {75.1} & 78.5 & {58.0} & {64.3} & {38.4} & 47.6 \\
\quad Li \etal \cite{li2018learning} & - & 84.9 & - & 77.2 & - & 66.8 & - & 47.7\\
\quad Huang \etal \cite{huang2020self} & \textcolor{black}{80.5}  & 85.2 & \textcolor{black}{75.4} & 77.1 & {58.4} & 65.5 & {39.4} & 46.6 \\
\quad Shi \etal \cite{shi2021gradient}& \textcolor{black}{82.0} & 85.5 & {76.9}& 77.8 & {62.0} & 68.6 & {40.2} & 45.1 \\ 
 \midrule
\multicolumn{9}{l}{\textit{Test-time domain generalization}}\\
\quad Wang \etal \cite{wang2021tent} 
& 83.9 & 85.2 & 72.9  & 73.0 & 60.9 & 66.3 & 33.7 & 37.1 \\
\quad Iwasawa \etal \cite{iwasawa2021test}
&81.7 & 85.3 & 76.5  & \textbf{80.0} & 57.0 & 68.3 & 41.6 & 47.0 \\
\quad Dubey \etal \cite{dubey2021adaptive}
& - & 84.1 & -  & 78.0 & - & 67.9 & -  & 47.3 \\
\quad Jang \etal \cite{jang2022test}
& {81.9} & 84.1 & {77.3}  & 77.6 & 63.7 & 68.6 & 42.6  & 47.4 \\
\quad Chen \etal \cite{chen2023improved} 
& {83.8} & - & 76.9 & - & 62.0 & - & 43.2 & - \\
\quad Liu \etal \cite{liu2024cross} 
& - & 86.4 & - & 76.3 & - & 68.1 & - & 48.2 \\
\quad Xiao \etal \cite{xiao2022learning} & \underline{84.1} & \underline{87.5} & \textcolor{gray}{\textbf{77.8}}  & \textcolor{gray}{{78.6}} & \textbf{66.0} & \textbf{71.0} & \textcolor{gray}{\underline{44.8}}  & \textcolor{gray}{\underline{48.4}} \\
 \toprule
\rowcolor{lightorange}
\quad {\textit{\textbf{This paper}}}
& \textbf{85.5} \scriptsize{$\pm$0.2}        & {\textbf{88.2}} \scriptsize{$\pm$0.3}        & \textbf{78.0} \scriptsize{$\pm$0.3}    & \underline{78.7} \scriptsize{$\pm$0.4} & \textbf{66.0} \scriptsize{$\pm$0.3}  & \textbf{71.2} \scriptsize{$\pm$0.4} & \textbf{46.5} \scriptsize{$\pm$0.3}   & \textbf{49.3} \scriptsize{$\pm$0.6} \\

\bottomrule
\vspace{-6mm}
\end{tabular}
}
\caption{
\textbf{Comparisons on input-level shift} using ResNet-18 and ResNet-50.
We report results for Test-Time adaptation methods by \cite{wang2021tent} and \cite{liang2020we} from \cite{jang2022test}. Our results are averaged over five runs. Gray numbers for \cite{xiao2022learning} based on our reimplementation. Our method consistently performs either as the top-performing (bold) or as the second-best (underlined). 
}
\label{table:all_datasets}
\end{table*}

\subsection{Results on different distribution shifts}

Due to adaptively generating the parameters of different layers in the model, GeneralizeFormer is able to handle various distribution shifts at different levels. We demonstrate this ability on the input level, output level, and feature level.

\noindent
\textbf{Input-level shift.} 
First, we conduct experiments on the common domain generalization setting, where the domain shifts between the source and target domains mainly exist in the input images such as different image styles. Thus, we treat these domain shifts as the input-level distribution shifts. 

We report the averaged results per dataset based on both ResNet-18 and ResNet-50 in Table~\ref{table:all_datasets}.
Our method outperforms the ERM baseline and the standard domain generalization methods on all datasets and backbones.
Moreover, compared with the test-time adaptation Tent and generalization methods \cite{iwasawa2021test, chen2023improved, xiao2022learning, jang2022test}, our method also achieves the best performance on all datasets based on ResNet-18, and top-2 performance based on ResNet-50, demonstrating the effectiveness of the method on the input-level shifts. 
\textcolor{black}{We also compare with generalization methods that only update the BN statistics at test time, without changing the parameters, e.g., alpha-BN \cite{you2021test} and MetaNorm \cite{du2020metanorm}. Based on ResNet-18, we perform comparably on PACS and outperform the two methods on Office-Home by 2.8\% and 2.5\%, respectively.}

\begin{table*}[t]
  \centering
  
\begin{subtable}[h]{0.62\textwidth}
\vspace{-3mm}

\resizebox{1\columnwidth}{!}{
\begin{tabular}{llllll}
\toprule
\textbf{Methods} & Photo & Art & Cartoon & Sketch & \textit{Mean} \\
 \midrule
ERM & 41.2  &   40.9 & 53.7 &46.2  &45.5  \\
Wang \etal \cite{wang2021tent} & 42.0  &  41.6 & 56.5 & 53.8 & 48.5 \\
Zhou \etal \cite{zhou2021domain} & 56.3 & 44.5 & 55.8 & 46.7 & 50.8 \\
Li \etal \cite{li2022uncertainty} & \textbf{60.2} & 45.0 & 54.4 & 49.2 & 52.2\\
Zhang \etal \cite{zhang2022exact} & 57.9 & \textbf{46.0} & 55.3 & 50.0 & 52.3 \\
 \rowcolor{lightorange}
{\textit{\textbf{This paper}}} & 50.0 \scriptsize{$\pm$0.2} & 45.7 \scriptsize{$\pm$0.3} & \textbf{61.2} \scriptsize{$\pm$0.28} & \textbf{55.5} \scriptsize{$\pm$0.5} & \textbf{53.1} \scriptsize{$\pm$0.31} \\
\bottomrule
\end{tabular}}
\caption{Output-level shift.}
\label{table:dist_shift}
\end{subtable}
\begin{subtable}[h]{0.293\textwidth}
\vspace{-3mm}

\resizebox{1\columnwidth}{!}{
\begin{tabular}{ll}
\toprule
\textbf{Methods} & \textbf{Accuracy} \\ 
\midrule
ERM & 69.3\\
Wang \etal \cite{wang2021tent} & 68.8 \\
Long \etal \cite{long2018conditional} & 70.7 \\
Ganin \etal \cite{ganin2016domain} & 72.1 \\
\rowcolor{lightorange}
\textit{\textbf{{\textit{\textbf{This paper}}}}} & \textbf{73.7}  \scriptsize{$\pm$0.7} \\ 
\bottomrule
\end{tabular}}
\caption{Feature-level shift}
\label{tab:feautre_level}
\end{subtable}
\vspace{-1em}
\caption{\textbf{Comparisons on output-level and feature-level shifts} for ResNet-18 on PACS. Our method achieves the best overall performance.}
\vspace{-4mm}
\end{table*}

\noindent
\textbf{Output-level shift.}
In addition to the input-level shifts in domain generalization datasets, we also conduct experiments with output-level distribution shifts. 
The output-level shifts are designed by introducing category shifts \cite{shen2022association, park2023test} in the domain generalization setting.
Specifically, we assign samples of different classes for different source domains during training.
The experiments are conducted on \textit{PACS}, where data from three source domains and seven classes are available during training. 
We select 3 of the 7 classes from the first source domain, the other 2 classes from the second source domain, and the remaining 2 classes from the last source domain as the training data.
\textcolor{black}{Therefore, the source domains have different label spaces (3, 2, 2 classes) from the target domain (entire label space with 7 classes), leading to spurious correlations between domains and classes during training, which does not exist at test time.
After training on the spurious correlated source domains, the model is then evaluated on the target domain with all 7 classes.}
Thus, there are both input-level shifts across domains and output-level shifts across categories at test time.

We compare to existing out-of-distribution generalization methods \cite{zhou2021mixstyle, zhang2022exact} and test-time adaptation methods like Tent under this setting. The results are in Table~\ref{table:dist_shift}.
Our method again achieves the best overall performance and outperforms the other alternatives on two of the four target domains.
Despite not observing all categories in each source domain, our method adaptively generates the specific layer parameters to jointly address the input and output shifts at test time.

\noindent
\textbf{Feature level shifts.}
To demonstrate the ability of the method on different levels of distribution shifts, apart from the input-level and output-level shifts, we further conduct experiments on the feature-level shifts.
Following \cite{garg2023rlsbench}, we conduct the experiments on Living-17 \cite{santurkar2020breeds}, where the source and target distributions consist of different subpopulations of the same superclasses. 
Since the input and label space are the same while the shifts are the features for prediction, we treat the distribution shift as the feature-level shift.
We report the results based on ResNet-18 in Table~\ref{tab:feautre_level} and compare our method with alternative methods. As shown in the table, our method again outperforms the other methods on feature-level shifts.

Overall, %
the results demonstrate that by adaptively generating the layer-wise model parameters for different test distributions, the proposed method is able to handle various distribution shifts well.

\begin{figure*}[t]
\centering 
\centerline{   
    \includegraphics[width=2\columnwidth]{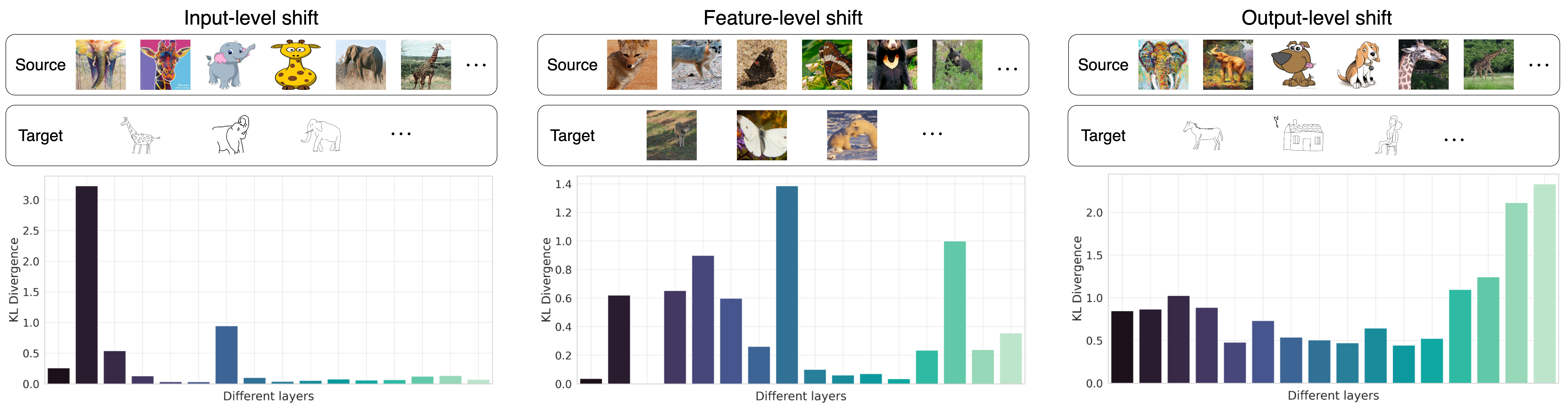}} 
    \vspace{-1em}
\caption{\textbf{Visualizations of adaptive model generation} \blue{using ResNet-18 on PACS. }(a) For input level shifts, based on the samples, our method focuses on generating the low-level layers. (b) Similarly, for feature-level shifts that consist of subpopulations, our method mainly changes the middle layers. (c) For the output level shifts, due to category shifts our method changes more on the high-level layers while also generating the initial layers since there are also input-level shifts in this setting. }
\vspace{-4mm}
\label{fig:visualization}  
\end{figure*}

\noindent
\textbf{Visualizations of adaptive model generation.} 
To further demonstrate the adaptive generation of the method for different distribution shifts, we also provide some visualizations of the differences between the source-trained model parameters and the generated ones. We calculate the layer-wise distance between the generated target-specific model parameters and the original source-trained one and compare the values with the other layers.
As shown in the first column in Fig.~\ref{fig:visualization}, the model generated across input-level shifts mainly changes the bottom layers. 
Other layers are also changed but not obviously.
For the feature-level distribution shifts (Fig.~\ref{fig:visualization} second column), the parameters in the middle layers change more obviously. 
In the last column, we find that the top layers generated by our method change more compared with the source-trained model. 
The low-level layers also show differences from the source model, which is because we also include input-shifts in this setting. 
The visualizations further demonstrate that our method adaptively generates the model parameters according to different distribution shifts.

\subsection{Ablation studies}
\label{sec: ablation}

\begin{table}[t]
\centering

\resizebox{0.95\columnwidth}{!}{
\begin{tabular}{lccll}
\toprule
& \multicolumn{2}{c}{\textbf{Generated layers}} &  \multicolumn{2}{c}{\textbf{Mean accuracy}} \\
\cmidrule(lr){2-3} \cmidrule(lr){4-5}
\textbf{Methods} & BN & Classifier & \textbf{PACS} & \textbf{Office-Home}\\
 \midrule
ERM Baseline & &  & 79.6 &  61.0 \\
 \midrule
\multirow{3}*{\textit{\textbf{This paper}}} & \ymark &  & \underline{85.3} \scriptsize{$\pm$0.2} &  64.8 \scriptsize{$\pm$0.3}\\
 & & \ymark & 84.0 \scriptsize{$\pm$0.2}& \underline{65.7} \scriptsize{$\pm$0.3} \\
 & \ymark & \ymark &  \textbf{85.5} \scriptsize{$\pm$0.2} &  \textbf{66.0} \scriptsize{$\pm$0.3}\\
\bottomrule
\end{tabular}}
\vspace{-2mm}
\caption{\textbf{Benefits of generating BN layers and classifiers} for ResNet-18 on \textit{PACS} and \textit{Office-Home}. Generating the parameters of both BN layers and classifiers improves the performance,  their combination achieves the best results. 
}
\label{tab:ab_fc}
\vspace{-6mm}
\end{table}

\noindent
\textbf{Benefits of generating different layers.}
We use GeneralizeFormer to generate the parameters of both Batch Normalization layers and classifiers for test-time domain generalization.
To show the benefits of both settings, we conduct experiments of generating the parameters of different layers separately.
We report results for a ResNet-18 on \textit{PACS} and \textit{Office-Home} in Table~\ref{tab:ab_fc}.
Generating BN and classifier parameters both improve the performance of the ERM baseline on the two datasets. 
Notably, generating BN parameters performs better than the classifiers on \textit{PACS} but worse on \textit{Office-Home}. 
The reason can be that \textit{Office-Home} contains more classes (65) than \textit{PACS} (7), which requires more adjustment on the classifier for generalization.
The results also demonstrate the benefits of generalizing different layers for different datasets or distribution shifts. 
Moreover, performance is further improved by adaptively generating both BN and classifier parameters, demonstrating the effectiveness of the proposed method. 
We report the results of adaptively generating both for the remaining ablations.

\begin{figure*}[t]
\centering 
\centerline{   
    \includegraphics[width=0.6\columnwidth]{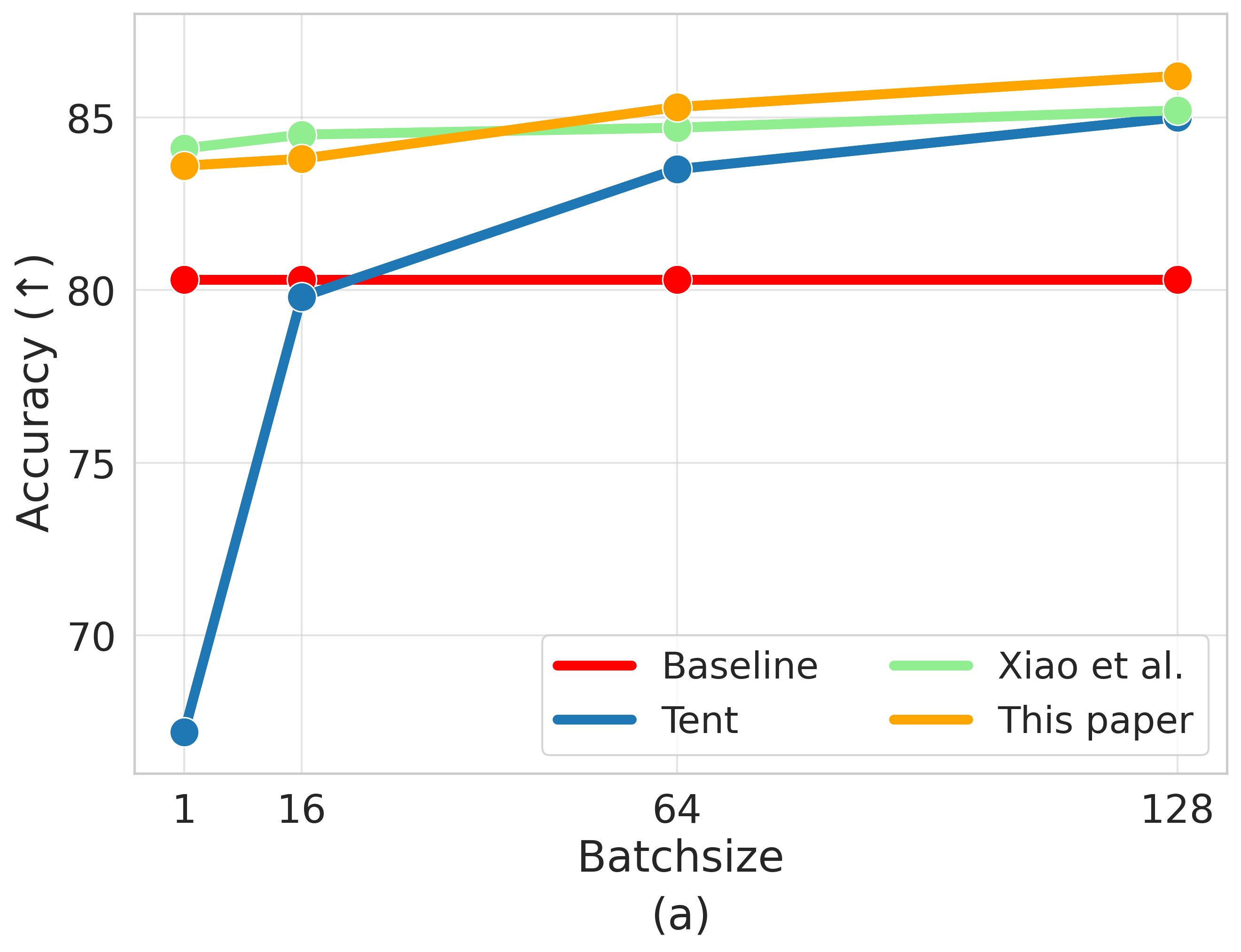} ~
    \includegraphics[width=0.66\columnwidth]{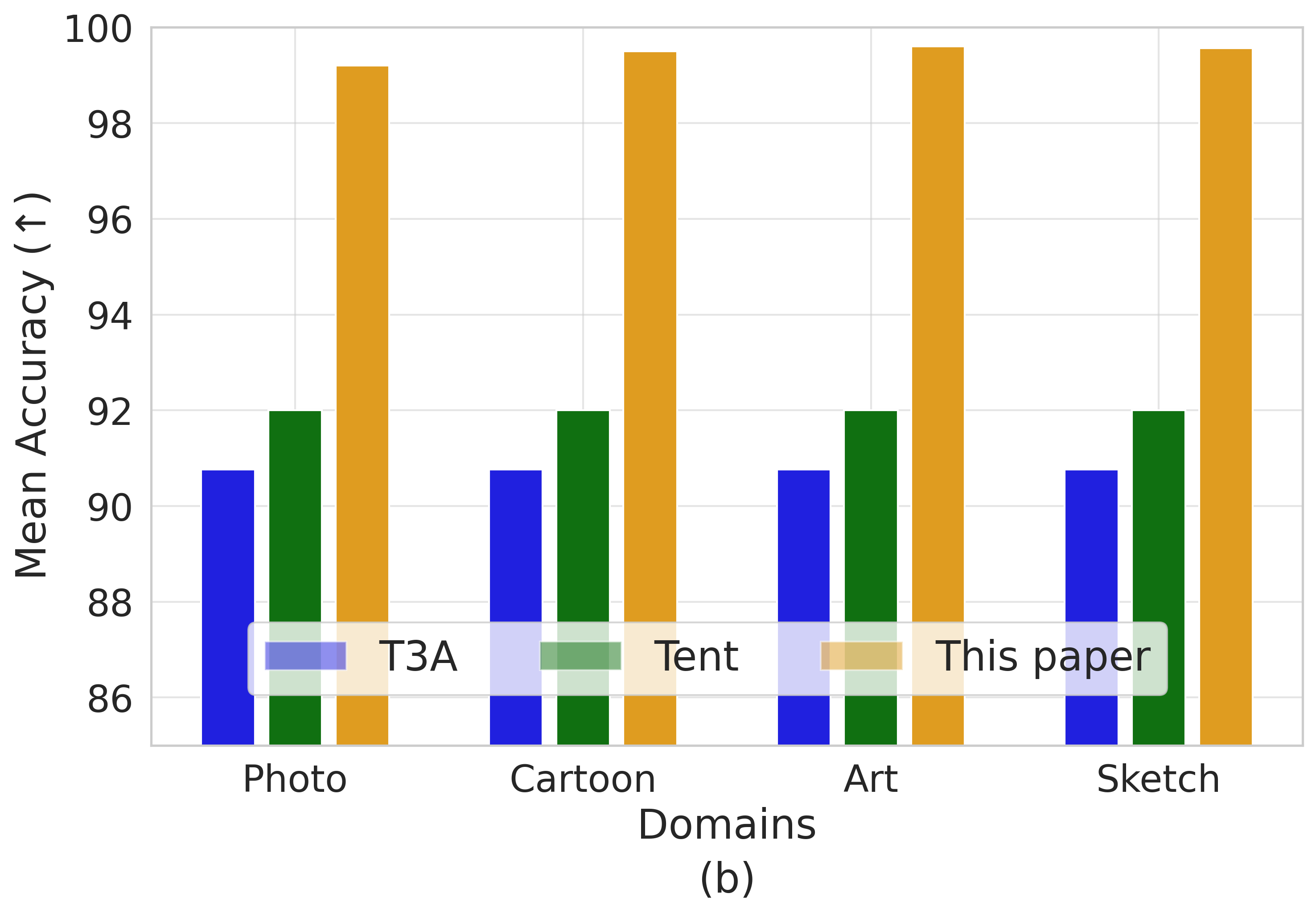} ~
    \includegraphics[width=0.66\columnwidth]{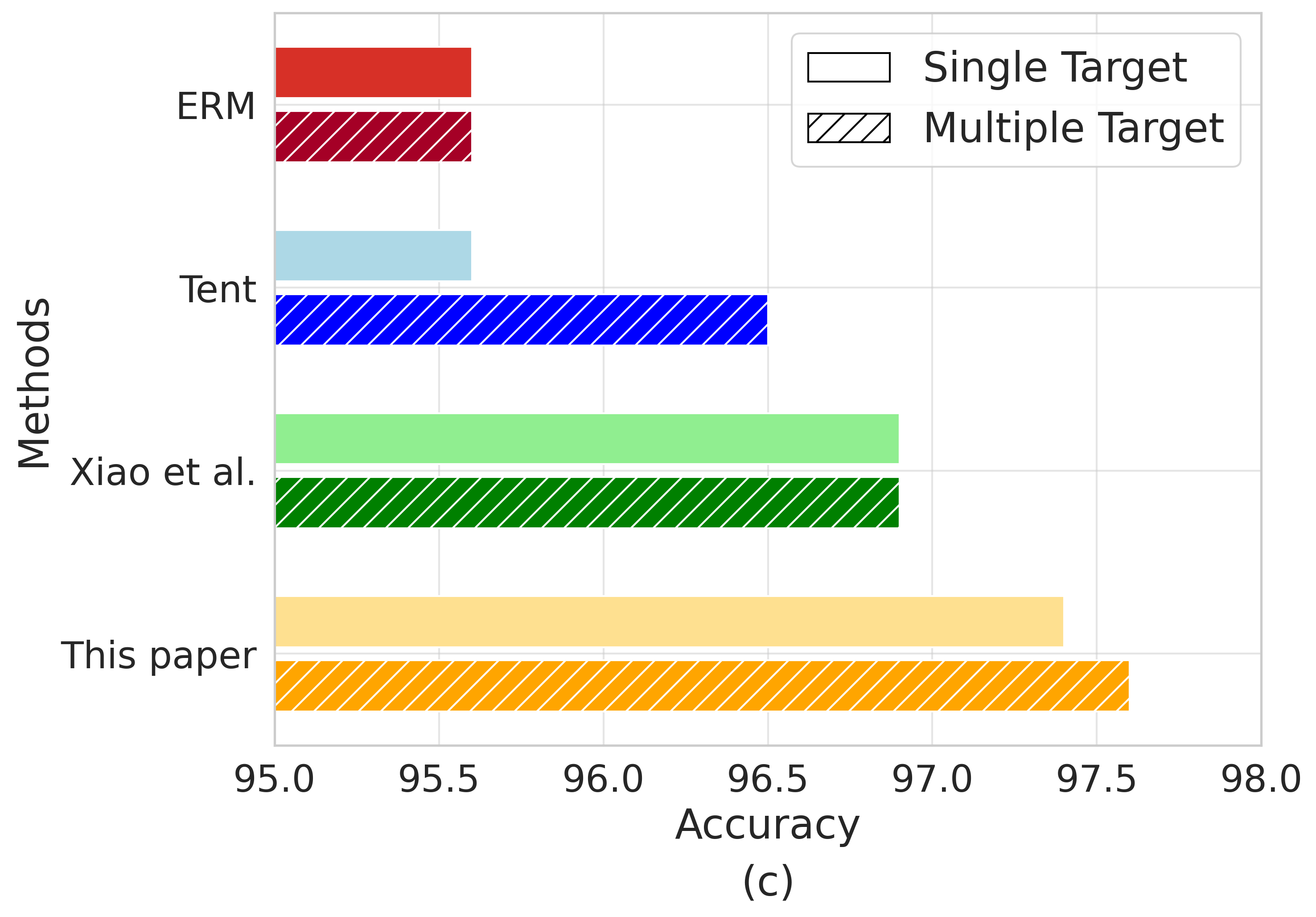} ~
} 
\vspace{-4mm}
\caption{\textbf{Generalization in different scenarios: (a) small batch sizes, (b) source forgetting, and (c) multiple target distributions.} Our method performs well in small batch sizes and complex scenarios with multiple distributions. The method also avoids source forgetting. 
}
\vspace{-4mm}
\label{fig:investigate-limdata_gf}  
\end{figure*} 

\vspace{0.5mm}
\noindent
\textbf{Generalization with small batch sizes.}
GeneralizeFormer learns the ability to adaptively generate the specific model for each target batch at test time. 
The requirement for batch sizes is more relaxed than the common fine-tuning and target statistics re-calculating methods, broadening the deployment in real applications. 
To demonstrate the effectiveness on small batch sizes, we compare with alternatives with different batch sizes. 
As shown in Fig.~\ref{fig:investigate-limdata_gf} (a), our method performs better than the common fine-tuning method Tent and the difference increases with small batch sizes. 
The proposed method can also handle the more challenging single-sample generalization problem with acceptable performance degradation while the performance of Tent collapses. Our performance on single samples is competitive with \cite{xiao2022learning}, which meta-learns the single-sample generalization ability.
Moreover, our method performs better with larger batch sizes while \cite{xiao2022learning} tends to saturate.
With the ability to generate sample-specific models, the proposed method can achieve generalization with very limited information and perform well in various complex scenarios.
We provide detailed results in the supplemental material.

\noindent
\textbf{GeneralizeFormer avoids source forgetting.}
Common fine-tuning methods update their model with successive steps to each batch of data, which can lead to catastrophic forgetting of source data. 
This problem is particularly detrimental during real-world deployment where the model is expected to maintain good performance on the source domains during generalization.
By contrast, GeneralizeFormer learns to directly generate target-specific model parameters for each test batch, so the information of the source distributions is not forgotten after \blue{adaptation}.
To demonstrate this, we conduct source-forgetting experiments on \textit{PACS}, \blue{where the model is re-evaluated on the source domains after adapting to the target domain.}
As shown in Fig.~\ref{fig:investigate-limdata_gf} (b), the common fine-tuning methods Tent and classifier adjustment method T3A \cite{iwasawa2021test} forget some knowledge learned from the source domains during the \blue{adaptation} at test-time, leading to performance decline on the source data. 
By contrast, our method directly generates the target-specific model parameters, without changing the knowledge in the source-trained model. \textcolor{black}{Hence, our method obtains similar performance (\{99.2\%, 99.6\%, 99.5\%, 99.6\%\}) as the source-trained models (\{99.7\%, 99.7\%, 99.8\%, 99.7\%\}) on four domains of PACS. 
}
This enhances the model's overall generalization performance in real-world deployment scenarios. 

\noindent
\textbf{Generalization for multiple target distributions.}
GeneralizeFormer adaptively generates model parameters for each batch of target data at test time. 
This enables the model to handle various target domains at test time, as is common in dynamic real-world scenarios. To show the benefits of the method in dynamic scenarios, we further conduct experiments with multiple target domains at test time on Rotated MNIST, where we use source domains with rotated angles $0^\circ$, $15^\circ$, $75^\circ$, and $90^\circ$, $30^\circ$, $45^\circ$, and $60^\circ$ as \blue{unseen target domains and unknown domain ids.} 
We report the results in both single-target and multiple-target settings in Fig.~\ref{fig:investigate-limdata_gf} (c).
In the single-target setting, the source-trained model is evaluated on each single target domain and the final accuracy is the average of the single-target results. 
In the multiple-target setting, the model is evaluated on all the target domains simultaneously, where the test samples would come from any target domain during test-time generalization.
As shown in Fig.~\ref{fig:investigate-limdata_gf} (c), the common fine-tuning method Tent performs well under the single-target setting. However, its performance under the multiple-target setting is worse, even comparable with the ERM baseline.
The reason is that the target samples of different domains contain conflicting information, which harms the online adjustment of the model.
Xiao \etal \cite{xiao2022learning} performs more robustly on the multiple-target setting since their model is trained to generalize on each target sample.
Nevertheless, our method achieves the best performance on both settings since we learn to adaptively generate model parameters according to the test-time target samples.

\begin{table}[t]
\begin{minipage}{0.26\textwidth}
\vspace{-2.5mm}
\resizebox{\columnwidth}{!}{
\begin{tabular}{lr}
\toprule
\textbf{Methods} & \textbf{Time} \\ 
\midrule
Jang \etal \cite{jang2022test} & 9m 30s \\
Liang \etal \cite{liang2020we} & 4m 22s  \\
Wang \etal \cite{wang2021tent} & 3m 16s \\
Iwasawa \etal \cite{iwasawa2021test} & 33s \\
\rowcolor{lightorange}
\textit{\textbf{This paper }} & 20s \\ 
\bottomrule
\end{tabular}}
\end{minipage} ~
\begin{minipage}{0.18\textwidth}
\caption{\textbf{The inference time for test generalization.} By replacing fine-tuning with model generation at test time, GeneralizeFormer requires minimal time for generalization.
}
\label{tab:inf_time_new}
\end{minipage}
\vspace{-6mm}
\end{table}

\noindent
\textbf{Inference time for generalization.} 
We provide the time cost of generalization on the target set at test time and compare it with other test-time methods.
The experiments are conducted on PACS based on ResNet-18.
As shown in Table~\ref{tab:inf_time_new}, GeneralizeFormer consumes less time than the fine-tuning-based methods Tent and \cite{liang2020we}. %
The proposed method achieves even slightly less time cost than the classifier adjustment method~\cite{iwasawa2021test}, which adjusts the classifier online at test time. 
\textcolor{black}{We also conduct experiments to generate all parameters of the model on PACS with a ResNet-18.  The method consumes 1m 10s, which is more computationally expensive than generating BN and classifier parameters (20s as shown in Table \ref{tab:inf_time_new}).}
We provide additional computational time comparisons in the supplemental materials.

\section{Conclusion}
\textit{GeneralizeFormer} learns to achieve test-time generalization with target information by adaptively generating model weights for varying target distributions. Under the introduced meta-generalization stage, the method learns to generate the target-specific model parameters by incorporating target information and source-trained weights through a transformer. By bypassing fine-tuning operations, we achieve both data and time-efficient generalization at test time with the target information, which also mitigates the instability and source forgetting caused by cyclical adaptation and prediction in conventional optimization of the source model. By its adaptive nature, the proposed method can handle various distribution shifts effectively and efficiently. 

\noindent Since we utilize meta-learning on multiple source domains to learn the generalization ability with more target information, the generalization of a single source domain is a limitation of the proposed method, which we consider a valuable avenue for future work. Especially, recent generative modeling techniques allow the creation of multiple source domains with varying shifts, which broadens the scope of our proposal.

\section*{Acknowledgments}
\noindent This work is financially supported by Core42, the University of Amsterdam, and the allowance Top consortia for Knowledge and Innovation (TKIs) from the Netherlands Ministry of Economic Affairs and Climate Policy.

{\small
\bibliographystyle{ieee_fullname}
\bibliography{main}
}

\newpage
\appendix

\section{Algorithms}
\label{alg:generalizeformer}
In this section, we provide the algorithms for source training and test-time generalization in Algorithm~\ref{alg:GF1} and \ref{alg:GF2}. 

\begin{algorithm}[ht!]
\small
\caption{Training for GeneralizeFormer \\
{\textbf{Input:}} $\mathcal{S}=\left \{ D_{s} \right \}^{S}_{s=1}$: source domains with corresponding ${(\x_s, \y_s})$; $\btheta$: model parameters of backbone; $\bphi$: model parameters of Transformer;
$\mathcal{B}_{tr}$: batch size during training; 
$N_{iter}$: the number of iterations. \\
{\textbf{Output:}}
Learned $\btheta, \bphi$
}
\label{alg:GF1}
\begin{algorithmic}[1]
\FOR{\textit{iter} in $N_{iter}$}
\STATE \textcolor{gray}{{Mimicking the domain shifts}} \\ $\mathcal{T'}$ $\leftarrow$ Randomly Sample ($\left \{ D_{s} \right \}^{S}_{s=1}$, $t'$); 
\\
$\mathcal{S'}$ $\leftarrow$ $\left \{ D_{s} \right \}^{S}_{s=1}$ $\backslash$ $\mathcal{T'}$; 

\STATE Sample datapoints $\{(\mathbf{x}_{s'}^{(k)}, \mathbf{y}_{s'}^{(k)})\}_{k=1}^{\mathcal{B}_{tr}} \sim \mathcal{S'}$, $\{(\mathbf{x}_{t'}^{(k)}, \mathbf{y}_{t'}^{(k)})\}_{k=1}^{\mathcal{B}_{tr}} \sim \mathcal{T'}$.
\STATE \textbf{Meta-source stage:} 
\\
\STATE 
Obtain meta-source model by training with the cross-entropy loss ($\mathcal{L}_{\mathrm{CE}}$) on meta-source labels and predictions 
$\btheta_{s'} = \mathop{\min}\limits_{\btheta} \mathbb{E}_{(\x_{s'}, \y_{s'}) \in \mathcal{S}'} [\mathcal{L}_{\mathrm{CE}}(\x_{s'}, \y_{s'}; \btheta)]$
\STATE \textbf{Meta-generalization stage:} 
\STATE Calculate meta-target features by $\z_{t'} {=} f_{\btheta_{s'}}(\x_{t'})$
\STATE Calculate layer-wise gradients with unsupervised loss function by $\g_{t'}^l {=} {\partial \mathcal{L} (\x_{t'})} / {\partial \btheta^l_{s'}}$

\STATE Generate the meta-target batch norm and classifier parameters of each layer by \\
$\btheta^l_{t'} {=} \bphi(\btheta^l_{s'}, \z_{s'}, \g_{s'}^l), \forall l=1,2,\cdots, L$, \\

\STATE Optimize transformer $\bphi$ by 

$\bphi = \mathop{\min}\limits_{\bphi}  \mathbb{E}_{(\x_{t'}, \y_{t'})}[\mathcal{L}_{\mathrm{CE}}(\x_{t'}, \y_{t'};\btheta_{t'})], \btheta_{t'} = \{ \btheta^l_{t'} \}_{l=1}^L$

\ENDFOR
\end{algorithmic}
\end{algorithm}

\begin{algorithm}[ht!]
\small
\caption{Test-time Generalization by GeneralizeFormer \\
{\textbf{Input:}} $\mathcal{T}$: target domain with $N_t$  unlabeled samples ${\x_t}$; $\btheta_s, \bphi_s$: source trained model parameters;
$\mathcal{B}_{te}$: batch size for each online step at test time.
}
\label{alg:GF2}
\begin{algorithmic}[1]
\FOR{\textit{iter} in $(N_t / \mathcal{B}_{te})$}
\STATE Sample one batch of target samples from the target domain $\{(\mathbf{x}_{t}^{(k)}\}_{k=1}^{\mathcal{B}_{te}} \sim \mathcal{T}$.

\STATE Calculate meta-target features by $\z_{t} {=} f_{\btheta_{s}}(\x_{t})$
\STATE Calculate layer-wise gradients with unsupervised loss function by $\g_{t}^l {=} {\partial \mathcal{L} (\x_{t})} / {\partial \btheta^l_{s}}$

\STATE Generate the meta-target batch norm and classifier parameters of each layer by \\
$\btheta^l_{t} {=} \bphi(\btheta^l_{s}, \z_{t}, \g_{t}^l), \forall l=1,2,\cdots, L$, \\

\STATE Make predictions by $p(\y|\x,\btheta_{t}), \btheta_{t} = \{ \btheta^l_{t} \}_{l=1}^L$
\ENDFOR
\end{algorithmic}
\end{algorithm}

\section{Additional Implementation details}
We follow the training setup as \cite{iwasawa2021test} that includes dataset splits and hyperparameter selection for our method. We utilize Imagenet pretrained ResNet-18 and ResNet-50 models for all domain generalization datasets, which is the same as previous methods. In the main paper, ERM baseline refers to evaluating the source-trained model directly on the given target set without any model adjustment at test time \cite{gulrajani2020search}. 

We describe the training and test-time procedures in the algorithm section. We implement the lightweight $\bphi$ model with the PyTorch transformer encoder module and utilize only one GPU to run the experiments on ResNet-18. {Following the common convention in the literature, e.g. [26, 32, 61], we utilize the given annotations of different domains that are predefined in the common domain generalization datasets. }The only hyperparameter involved here is the number of layers that we have experimented with in Section~\ref{section:additional_results}. We utilized identical settings and hyperparameters in the main paper for all domain generalization benchmarks. We utilize the train domain validation selection method to obtain the model for test-time domain generalization same as \cite{iwasawa2021test}. 

For test-time generalization, we utilize a small batch of 20 samples per batch. We generate the target model parameters using the $\bphi$ model and do not perform any backpropagation on the source model, which helps in reducing our computational time, as shown in Section 4 of the main paper. We do not have any additional hyperparameters in our method and utilize PyTorch to implement the method. For ResNet-18 models, we require only one GPU and utilize NVIDIA 1080Ti. We conducted all the experiments using five different random seeds. We will release the code in the final version as a link to the public repository. \\

\noindent {\textbf{Architecture of Generalizeformer. }GeneralizeFormer utilizes the transformer-encoder module from Pytorch with 8 layers, which consists of multi-head attention modules and feedforward modules. 
The source parameters, target features, and gradients are all formatted to the same dimension as the source parameters (e.g., 512 for the last block for ResNet-18) and then concatenated and utilized as input tokens to the transformer. 
The attention is calculated between these inputs to enhance each other according to their relationships.
We use the output features of the source parameters as the generated target parameters, which therefore match the dimensions of the source ones. At inference, we directly replace the source parameters with the generated parameters (Section 3 of the main paper)}. \\

\noindent \textbf{Runtime Comparison. } To show the efficiency of our method, we provide comparisons of the runtime cost at both training and test time, as well as the memory usage during training.
Our method requires more training time and parameters while having the lowest time cost at test time compared with other test-time generalization methods. 
Due to the meta-generalization stage, the proposed method takes 9 hours for 10,000 iterations utilizing a ResNet-18 and an NVIDIA Tesla 1080Ti GPU, which is longer than the ERM baseline of 6.5 hours.
In addition, the ERM method based on ResNet-18 requires 11.18M memory. In contrast, our method requires 39M when utilizing the 8-layer transformer for parameter generation. 
Our method can also be implemented with 4-layer and 2-layer transformers, which consume 32\% and 48\% fewer memory with similar performance of 85.2\% and 84.9\% on PACS, respectively.

Moreover, we also provide the computational time comparisons during test-time generalization on different datasets (Table~\ref{table:runtime_ttdg}), which is more important for test-time methods. 
We consume less time than all alternative methods in Table \ref{table:runtime_ttdg} on four domain generalization datasets. This ability is ideal for real-world deployment scenarios. The proposed method even consumes less time on than the classifier adjustment method \cite{iwasawa2021test} that only updates the classifier.

\noindent\textbf{Datasets details. } As mentioned in the main paper, we perform the experiments on image classification problems and demonstrate its effectiveness on six datasets namely: \textit{PACS} \cite{li2017deeper}, \textit{VLCS} \cite{fang2013video}, \textit{Office-Home}\cite{venkateswara2017deep}, \textit{TerraIncognita} \cite{beery2018recognition}, \textit{Living-17} \cite{santurkar2020breeds}, \textit{Rotated MNIST} and \textit{Fashion MNIST} \cite{piratla2020efficient}. 
\textit{PACS} \cite{li2017deeper} consists of 9,991 samples, 7 classes, and 4 domains: Photo, Art-painting, Cartoon, and Sketch. 
\textit{VLCS} \cite{fang2013video} consists of 10,729 samples, 5 classes and 4 domains: Pascal, LabelMe, Caltech, and SUN. 
\textit{Office-Home} \cite{venkateswara2017deep} consists of 15,5000 images, 65 classes and 4 domains: Art, Clipart, Product, and Real-World. 
\textit{TerraIncognita} \cite{beery2018recognition} consists of 34,778 samples, 65 classes, and 4 domains:  Location 100, Location 38, Location 43, and Location 46. We followed \cite{li2017deeper} for training and validation split. 
We follow the ‘leave-one-out” protocol \cite{li2017deeper,carlucci2019domain} by evaluating the model on each target domain with the parameters trained on the other source domains. 
We utilize \textit{Living-17} \cite{santurkar2020breeds}, which contains 17 classes with subclasses and 39780 images in source while 1700 images in target. Our performance is reported on the target domain.
For MNIST and Fashion-MNIST, we utilize the \textit{rotated MNIST} and \textit{rotated Fashion-MNIST} and follow \cite{piratla2020efficient} where the images are rotated by different angles for different domains. 
We use the subsets with rotation angles from $15^\circ$ to $75^\circ$ in intervals of $15^\circ$ as five source domains, and images rotated by $0^\circ$ and $90^\circ$ as the target domains.

\begin{table}[ht!]
\centering

\resizebox{0.9\columnwidth}{!}{%
		\setlength\tabcolsep{4pt} 
\begin{tabular}{lllll}
\toprule

 & VLCS & PACS & Terra & OfficeHome  \\ \midrule
 Tent \cite{wang2021tent}   & 7m 28s & 3m 16s & 10m 34s & 7m 25s \\
 Tent \cite{wang2021tent} (BN)   & 2m 8s & 33s  & 2m 58s & 1m 57s\\
 SHOT \cite{liang2020we} & 8m 09s & 4m 22s  & 12m 40s &8m 38s \\
 TAST \cite{jang2022test} & 10m 34s & 9m 30s  & 26m 14s & 22m 24s \\
 T3A \cite{iwasawa2021test}    & 2m 09s & 33s & 2m 59s & 2m 15s \\
\textit{\textbf{This paper}}  & 47s & 20s & 52s & 44s  \\

\bottomrule

\end{tabular}
}
\caption{{\textbf{Computational time comparison on different datasets with ResNet-18 as a backbone network during test-time generalization.}} 
The proposed method has better overall time consumption than existing test-time adaptation and test-time domain generalization methods.
}
\label{table:runtime_ttdg}
\end{table}

\section{Additional results and discussion}
\label{section:additional_results}

\noindent {\textbf{Why GeneralizeFormer works. }}
{To achieve good performance in a target domain, obtaining target-specific model parameters is crucial. Existing fine-tuning methods approximate target parameters by MAP estimation with an unsupervised loss (Section 3 of the main paper). Since their approximation depends on the original parameter quality and the number of target samples, errors accumulate. Our method avoids this by directly inferring batch-specific parameter distributions for each target batch in a feedforward pass. By doing so, our method is more practical for scenarios where the number of test samples is small, the test tasks are unknown, and a specific model cannot be selected, as evident in Figure 4 of the main paper. } \\

\noindent \textcolor{black}{{\textbf{How Generalizeformer retains source data.}} \textcolor{black}{The motivation behind the method is online adaptation can lead to error accumulation and forgetting due to iterative backpropagations. To address this issue, we learn a transformer to directly generate the parameters for each target batch individually. Therefore, the generated parameters are specific to each target sample, without affecting other batches. The source-specific parameters can also be recalled by inferring model parameters using each source batch, therefore avoiding source forgetting. We have added this discussion to the appendix.}}  \\ \\
\textbf{Further clarification of adaptively generating layer-wise model parameters per sample.  }
Technically, we introduce a transformer for parameter generation, whose attention mechanism effectively aggregates useful knowledge in source parameters and target features to avoid information loss.
We further consider layer-wise gradients per target batch as input of the transformer, which indicates the relationships between each layer of the source parameters and each target batch. By doing so, the gradients guide model generation for each layer and for different target domains, batches, and even samples.
This reduces error transmission among target samples and layers while enhancing the generalization ability across samples and domain shifts. Meta-learning is utilized just to mimic domain shifts to learn the ability of model generation; we do not claim it as a contribution.\\

\begin{figure}[t]
    \centering
    \begin{subfigure}[b]{0.45\linewidth}
        \includegraphics[width=\linewidth]{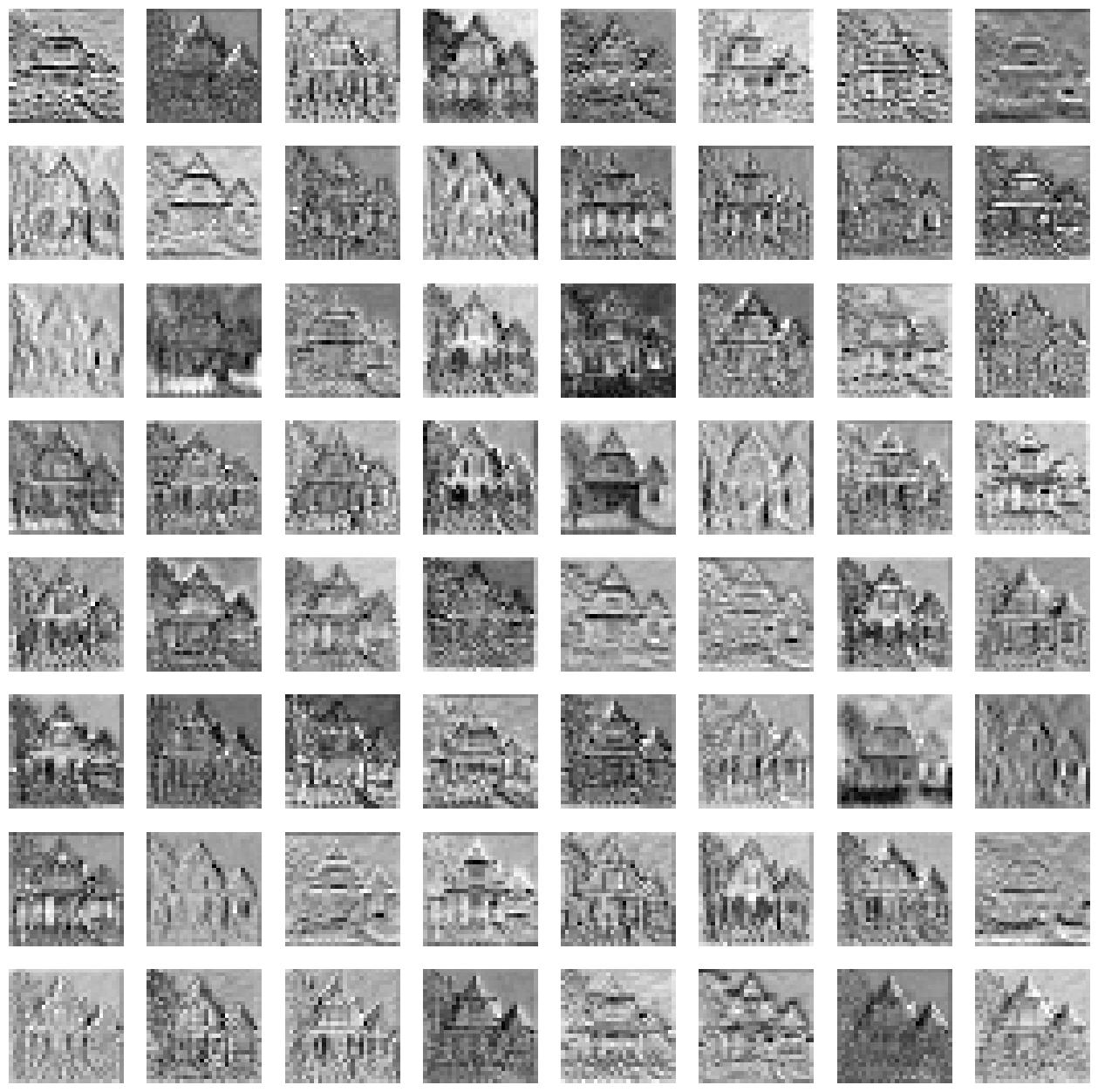}
        \caption{}
        \label{fig:weightsgen}
    \end{subfigure}
    \hfill
    \begin{subfigure}[b]{0.45\linewidth}
        \includegraphics[width=\linewidth]{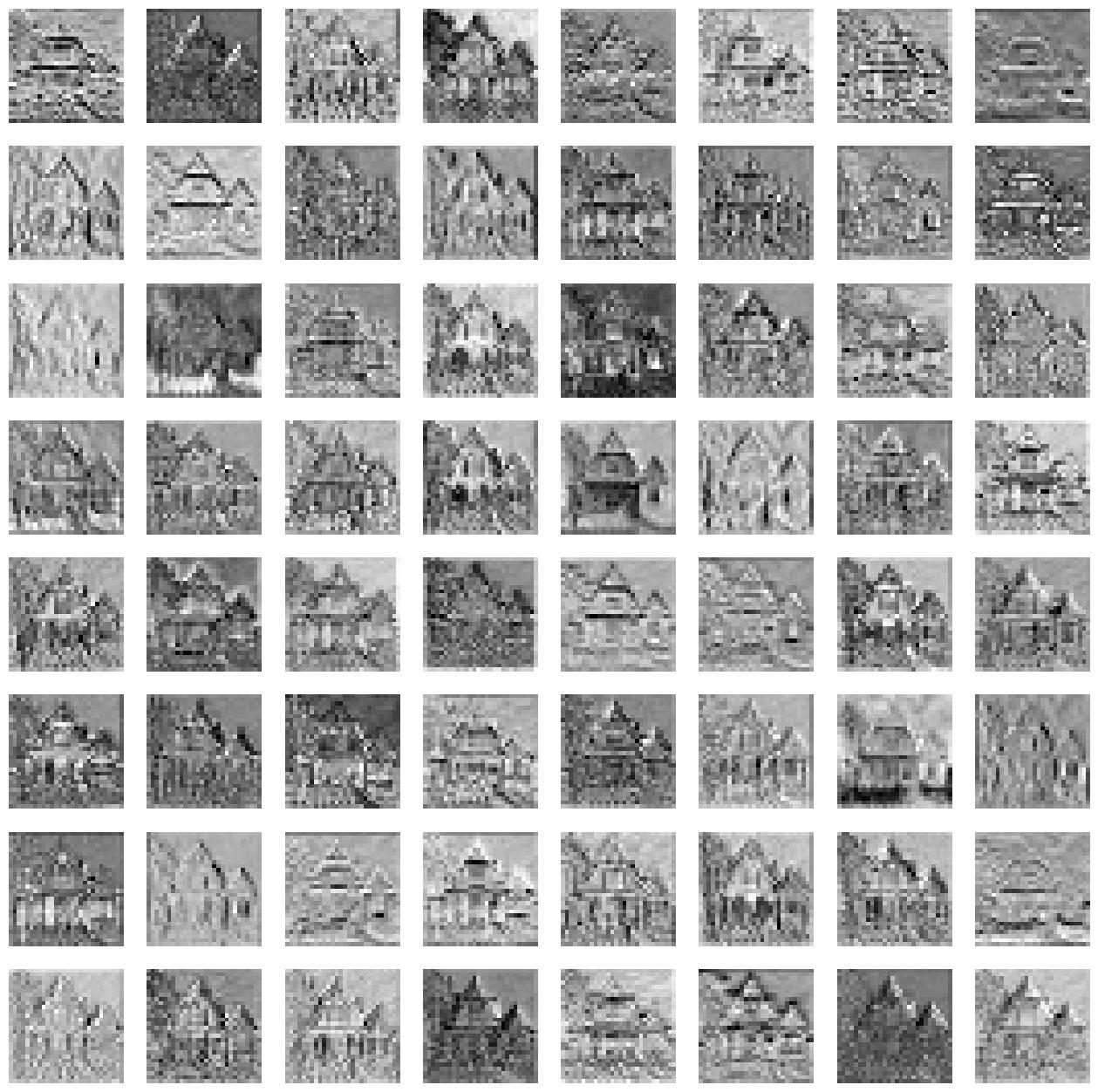}
        \caption{}
        \label{fig:realweights}
    \end{subfigure}
    \caption{{\textbf{Visualization of generated weights} on PACS. Each row visualizes a 28x28 filter from the batch norm layer for a sample image from the photo domain. We show the (a) Generated weights by GeneralizeFormer (b) Real weights.}}
    \vspace{-6mm}
    \label{fig:generated_weights}
\end{figure}

\noindent \textbf{Visualization of generated weights.}
In Fig.~\ref{fig:generated_weights}, we provide a visualization of the generated weights and real weights through filters for the photo domain PACS dataset. The filters obtained through the use of (a) generated weights are identical to the filters obtained through the use of (b) real weights. \\

\begin{figure}[t]
    \vspace{5mm}
    \centering
    \includegraphics[width=0.90\linewidth]{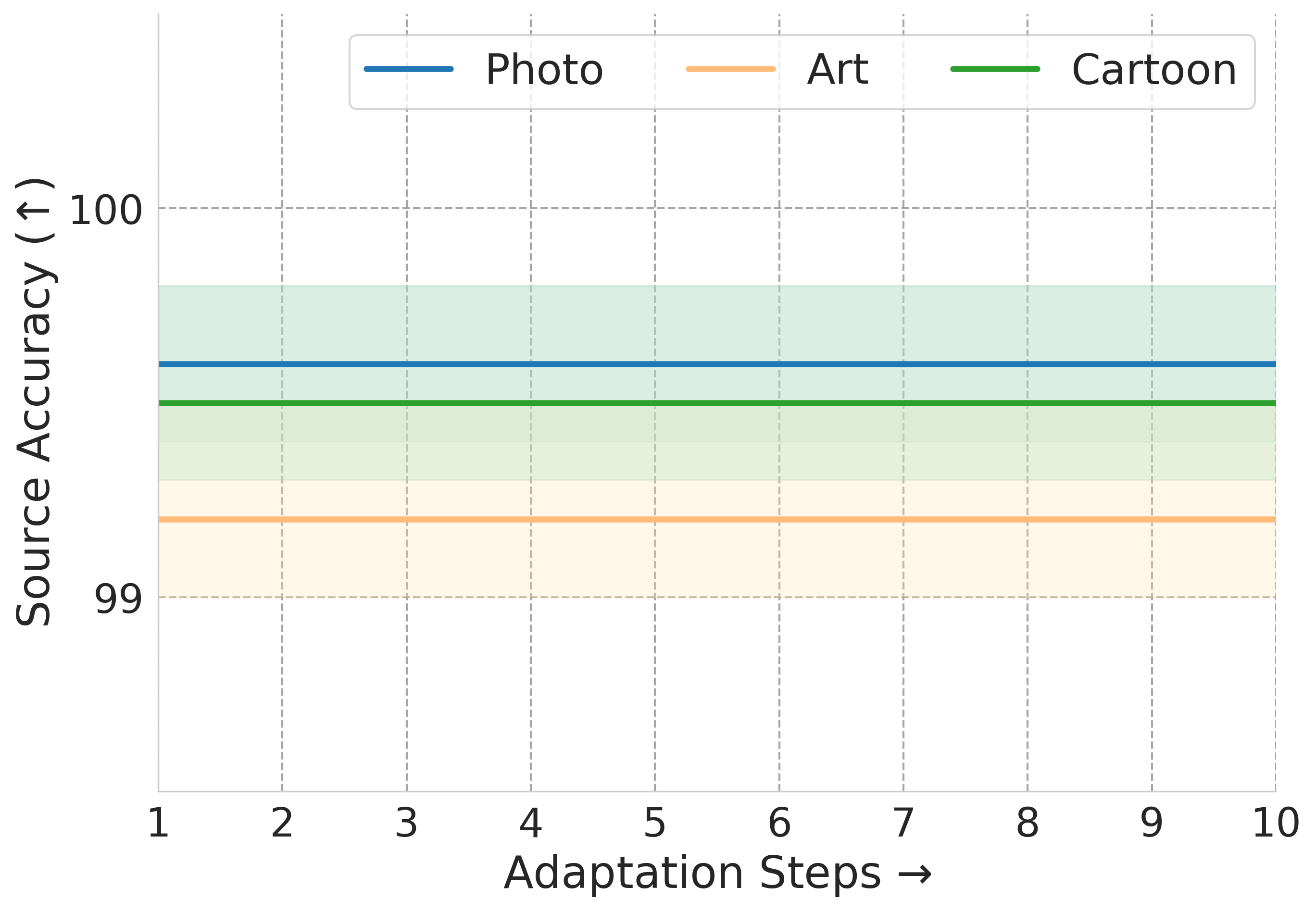}
    \caption{\textcolor{black}{\textbf{Avoiding source forgetting across adaptation steps of our method} with ResNet-18. Each line graph represents the accuracy on the source domains by utilizing the model which was adapted on the sketch domain of the PACS dataset.  }}
    \vspace{-2mm}
    \label{fig:source_forget}
    
\end{figure}

\noindent\textcolor{black}{\noindent\textbf{Avoiding source forgetting across steps. }
In Fig.~\ref{fig:source_forget}, we also provide the visualization of retaining the source information across adaptation steps. For this experiment, initially, at test-time, the model is adapted to the sketch domain of the PACS dataset. Next, the adapted model is re-evaluated on the source domains: photo, art-painting, and cartoon to evaluate the performance on the source domains. The conclusion is similar to the ablation study of avoiding source forgetting from the main paper, where our method retains the source data.}

\noindent\textbf{Ablation of different inputs for $\bphi$ network.} 
As aforementioned in the methodology section, the $\bphi$ model utilizes the target features, source-trained parameters and layer gradients to generate the target parameters for test-time generalization. In Table~\ref{tab:ablate_inputs}, to show the benefits of utilizing these three inputs, we perform an ablation study by utilizing a subset of the inputs in each experiment. Notably, all inputs help in achieving the best performance. 
The source parameters provide the basic ability of feature extraction and classification learned during training. Without it, it is difficult for fast model generation in one feedforward pass (82.0\% on PACS). The target features are essential for tailoring the generated model to specific target data, otherwise, it will cause unfitness (82.7\%). Without the gradients, it is difficult to adaptively control the generation of parameters, leading to performance degradation (81.9\%). The integration of all these inputs results in a comprehensive approach, leading to an improvement of 85.5\% to effectively and adaptively generate the target-specific parameters.\\

\begin{table}[t]
\centering
\resizebox{\columnwidth}{!}{
\begin{tabular}{lccll}
\toprule
& \multicolumn{3}{c}{\textbf{Inputs}} &  \multicolumn{1}{c}{\textbf{}} \\
\cmidrule(lr){2-4} 
 & Target features & Layer gradients & Source parameters & {Mean}\\
 \midrule
ERM Baseline & &  & & 79.6 \\
 \midrule
\multirow{3}*{\textit{\textbf{This paper}}} & \ymark & \ymark  &  &  82.0\scriptsize{$\pm$0.3}\\

 &  & \ymark & \ymark  & \underline{82.7}\scriptsize{$\pm$0.3} \\
 & \ymark &  & \ymark  &  {81.9}\scriptsize{$\pm$0.3}\\

 & \ymark & \ymark & \ymark  & \textbf{85.5}\scriptsize{$\pm$0.2} \\
\bottomrule
\end{tabular}}
\vspace{1mm}
\caption{\textbf{Ablation of different inputs for $\bphi$ network} for ResNet-18 on \textit{PACS}. Utilizing all three inputs achieves the best results, followed by using the layer gradients and source parameters.
}
\label{tab:ablate_inputs}
\vspace{-3mm}
\end{table}

\noindent\textbf{Analyses for only generating parameters of Batch Normalization layers and classifiers.}
We generate only the BN and classifier parameters for computational efficiency since they are low dimensional with much fewer parameters.
Moreover, BN and classifiers have significant influences on domain shifts.
BN parameters affect the statistics of the features, which contain style or domain information \cite{huang2017arbitrary}. Previous methods like Tent also update BN parameters to handle domain shifts. Additionally, classifier parameters further handle the domain shifts at the semantic-level, as also evident in T3A~\cite{iwasawa2021test} and Xiao \etal \cite{xiao2022learning}. Overall, by generating BN and classifier parameters, we handle domain shifts across different feature levels in an efficient way.\\

\noindent {\textbf{Generating Batch norms at different levels. }
From Table 3 in the main paper, generating only the classifier achieves 84.0\% on PACS.
We also conduct experiments to generate the BN layer in different blocks, where we get 84.5\%, 84.8\%, and 84.9\% for generating blocks 5, 6, and 7, respectively. 
All these settings underperform GeneralizeFormer (85.5\%), showing the effectiveness of adaptive generation of different layers.}\\

\noindent\textbf{Performance without meta-learning.} 
We also investigate the effectiveness of the meta-learning strategy in the proposed method on PACS.
Without meta-learning, the performance with ResNet-18 degrades from 85.5\% to 84.7\%, while still performing better than ERM (79.6\%) and other baselines.\\

\noindent\textbf{Models without Batch Normalization layers.}
We generate parameters of both normalization and linear layers, where the former seems to be more important in ResNet-based models.
However, next to affine parameters of Batch Normalization layers, the generation of the linear layer also performs well, achieving 84.0\% and 65.7\% on PACS and Office-Home from Table 4 in the main paper. 
This indicates that the proposed method can also be extended to handle domain shifts within other model architectures without Batch Normalization layers, e.g., MLP-based models or Transformers akin to T3A \cite{iwasawa2021test}.\\

\noindent\textbf{Detailed results of limited batch sizes.} As shown in the main paper, we conducted experiments using limited batch sizes. We also performed the challenging single-sample generalization setting that widens its scope for deployment in real applications. In Table~\ref{table:ablate_batch_size}, we provide detailed results of small batch sizes ablation from the main paper. The conclusion is similar to the main paper, where our method performs better than Tent \cite{wang2021tent}, and the difference increases with batch sizes. For single sample, we are competitive to \cite{xiao2022learning} while achieving better performance than it for larger batch sizes. By generating sample-specific models, the proposed method can achieve generalization with limited information.\\

\begin{table}[t]

\centering
	\resizebox{\columnwidth}{!}{%
		\setlength\tabcolsep{8pt}
\begin{tabular}{llllll}
\toprule
~ & \textbf{Photo}    & \textbf{Art}         & \textbf{Cartoon}       & \textbf{Sketch}        & \textit{Mean}       \\ \midrule

Baseline & 94.1 & 78.0 & 73.1 & 73.3 & 79.6\scriptsize{$\pm$0.4} \\
\rowcolor{mColor2}
\hline
\multicolumn{6}{l}{\textit{Test batch size = 1}} \\ 

\quad Tent \cite{wang2021tent} & 84.6  & 65.1& 69.5 & 49.7& 67.2\scriptsize{$\pm$0.4}\\
\quad Xiao \etal \cite{xiao2022learning} &95.8  & 82.0& 79.7 & 78.9 & 84.1\scriptsize{$\pm$0.2}\\
\quad \textbf{\textit{This paper}} & 95.5 &83.4  &80.4 &74.9 & 83.6\scriptsize{$\pm$0.2} \\
\hline
\rowcolor{mColor2}
\multicolumn{6}{l}{\textit{Test batch size = 16}} \\ 
\quad Tent \cite{wang2021tent} & 93.6   & 80.2& 76.9 & 68.4 & 79.8\scriptsize{$\pm$0.3}\\
\quad Xiao \etal \cite{xiao2022learning} &96.1  & 82.3 & 80.8 & 78.6 & 84.5\scriptsize{$\pm$0.2}\\
\quad \textbf{\textit{This paper}} & 96.4&82.0  &82.7 &74.0 & 83.8\scriptsize{$\pm$0.2} \\
\rowcolor{mColor2}
\hline
\multicolumn{6}{l}{\textit{Test batch size = 64}} \\ 
\quad Tent \cite{wang2021tent} & 96.0   & 81.9 & 80.3 & 75.9  & 83.5\scriptsize{$\pm$0.4} \\
\quad Xiao \etal \cite{xiao2022learning} &96.0  & 82.5 & 81.3 & 78.8 & 84.7\scriptsize{$\pm$0.2}\\

\quad \textbf{\textit{This paper}} &96.8 & 84.5& 83.6& 76.3 &85.3\scriptsize{$\pm$0.2}\\
\hline
\rowcolor{mColor2}
\multicolumn{6}{l}{\textit{Test batch size = 128}} \\ 
\quad Tent \cite{wang2021tent} & 97.2  & 84.9& 81.1& 76.8 & 85.0\scriptsize{$\pm$0.5}\\

\quad Xiao \etal \cite{xiao2022learning} &96.2  & 83.2 & 82.3 & 79.0 & 85.2\scriptsize{$\pm$0.2}\\

\quad \textbf{\textit{This paper}} &97.1 &85.7 & 85.2 &  76.9 & 86.2\scriptsize{$\pm$0.2} \\
\bottomrule
\end{tabular}
}
\caption{\textbf{Detailed results of limited batch sizes.} GeneralizeFormer performs better than Tent \cite{wang2021tent} with different batch sizes. The proposed method achieves competitive results with \cite{xiao2022learning} for small batch sizes and outperforms it on larger batch sizes. 
} 
\label{table:ablate_batch_size}

\end{table}

\noindent\textbf{Different losses for gradient information.}
As mentioned, the method can utilize different unsupervised losses for gradient information. In Table~\ref{table:ab_loss}, we utilize different unsupervised based losses such as \cite{zhang2021memo} and loss through pseudo labeling. Notably, unsupervised entropy minimization, which is the default loss function, performs well. This study shows the versatility of the proposed method, such that it can integrate different losses. Consequently, the efficacy and applicability of the method may be further improved by utilizing different unsupervised loss functions in the future.

\begin{table}[h]
\centering
\vspace{-1mm}
\resizebox{\columnwidth}{!}{
\begin{tabular}{lccccl}
\toprule
Strategies & Photo & Art & Cartoon & Sketch & Mean\\
 \midrule
 Memo \cite{zhang2021memo} &  96.2 & 82.1 & 81.5 & 70.0 & 82.5\scriptsize{$\pm$0.4} \\

 Pseudo labels   & 96.6 &  80.4 & 82.7 & 75.2  & 83.7\scriptsize{$\pm$0.3} \\

 Entropy Minimization &  96.9 & 85.0 & 83.3 & 76.7 & 85.5\scriptsize{$\pm$0.2} \\

\bottomrule
\end{tabular}}
\vspace{1mm}
\caption{\textbf{Different losses for gradient information} for ResNet-18 on PACS dataset. The proposed method can make use of different losses for the gradient information to achieve good performance. We utilize entropy minimization as the default for our experiments.}
\label{table:ab_loss}
\vspace{-4mm}
\end{table}

% \newpage
% {\small
% \bibliographystyle{ieee_fullname}
% \bibliography{main}
% }

% \end{document}

% \input{tex_files/sections/appendix}

\end{document}